\documentclass[sigconf]{acmart}

\usepackage{enumitem}
\usepackage{balance}
\usepackage{url}
\usepackage{amsmath}
\usepackage{mathtools,color}
\usepackage{graphicx}
\usepackage{multirow}
\usepackage[utf8]{inputenc}
\newcommand{\D}{\mathcal{D}}
\newcommand{\X}{\mathbf{X}}

\AtBeginDocument{%
  \providecommand\BibTeX{{%
    \normalfont B\kern-0.5em{\scshape i\kern-0.25em b}\kern-0.8em\TeX}}}

\copyrightyear{2023}
\acmYear{2023}
\setcopyright{acmlicensed}
\acmConference[SIGIR '23] {Proceedings of the 46th International ACM SIGIR Conference on Research and Development in Information Retrieval}{July 23--27, 2023}{Taipei, Taiwan.}
\acmBooktitle{Proceedings of the 46th International ACM SIGIR Conference on Research and Development in Information Retrieval (SIGIR '23), July 23--27, 2023, Taipei, Taiwan}
\acmPrice{15.00}
\acmISBN{978-1-4503-9408-6/23/07}
\acmDOI{10.1145/3539618.3591976}

\begin{document}

\title{Edge-cloud Collaborative Learning with Federated and Centralized Features}




\author{Zexi Li$^{\ast}$}
\affiliation{%
  \institution{Zhejiang University}
  \city{Hangzhou}
  \country{China}
}
\email{zexi.li@zju.edu.cn}

\author{Qunwei Li$^{\ast}$}
\affiliation{%
  \institution{Ant Group}
  \city{Hangzhou}
  \country{China}
}
\email{qunwei.qw@antgroup.com}

\author{Zhou Yi}
\affiliation{%
  \institution{The University of Utah}
  \city{Salt Lake City}
  \country{USA}
}
\email{yi.zhou@utah.edu}

\author{Wenliang Zhong$^{\star}$}
\affiliation{%
  \institution{Ant Group}
  \city{Hangzhou}
  \country{China}
}
\email{yice.zwl@antgroup.com}

\author{Guannan Zhang}
\affiliation{%
  \institution{Ant Group}
  \city{Hangzhou}
  \country{China}
}
\email{zgn138592@antgroup.com}

\author{Chao Wu$^{\star}$}
\affiliation{%
  \institution{Zhejiang University}
  \city{Hangzhou}
  \country{China}
}
\email{chao.wu@zju.edu.cn}

\thanks{$^\ast$ Both authors contributed equally to this work.} 
\thanks{$^\star$ Corresponding authors.}
\thanks{Work was done when Zexi Li was an intern at Ant Group.}





\renewcommand{\shortauthors}{Li and Li, et al.}
\renewcommand{\authors}{Zexi Li, Qunwei Li, Yi Zhou, Wenliang Zhong, Guannan Zhang, Chao Wu}
\begin{abstract}
Federated learning (FL) is a popular way of edge computing that doesn't compromise users' privacy. Current FL paradigms assume that data only resides on the edge, while cloud servers only perform model averaging. However, in real-life situations such as recommender systems, the cloud server has the ability to store historical and interactive features. In this paper, our proposed \textbf{E}dge-\textbf{C}loud \textbf{C}ollaborative Knowledge \textbf{T}ransfer Framework (\textbf{ECCT}) bridges the gap between the edge and cloud, enabling bi-directional knowledge transfer between both, sharing feature embeddings and prediction logits. ECCT consolidates various benefits, including enhancing personalization, enabling model heterogeneity, tolerating training asynchronization, and relieving communication burdens. Extensive experiments on public and industrial datasets demonstrate ECCT's effectiveness and potential for use in academia and industry.
\end{abstract}





\maketitle

\section{Introduction}
In the conventional centralized training paradigm, data is collected from multi-sourced edge devices, based on which a large model is trained at a cloud server \cite{liao2009machine,salman2018deepconf}. Recently, due to user privacy and communication efficiency concerns, federated learning (FL) has been proposed as an alternative to the centralized training paradigm \cite{li2022can,mcmahan2017communication,DBLP:journals/spm/LiSTS20,wang2021field,li2022mining}. FL assumes all the training data is kept locally at the edge devices, and a global model can be obtained by iterative edge training and cloud-side model averaging. It has shown feasibility and applicability in privacy-preserving scenarios. For example, keyboard inputs are sensitive information of cellphone users, and Google has used FL to improve next-word prediction while preserving users' privacy \cite{bonawitz2019towards}.

However, the current FL frameworks ignore the existence of cloud-side data, and it does not directly incorporate cloud training. In industrial applications, the cloud server usually has rich computation resources to train a big model with rich historical and non-private features. Taking the recommender system (RS) as an example, privacy-sensitive features, including one user's gender, age, and other real-time features such as the user's timely interactions with the items are generally inaccessible to the cloud and are stored locally in the devices. We denote these features as \textbf{"federated features"}. On the other hand, the centralized historical features (with fewer privacy concerns), including the user's historical interactions with items, item categories, item flags, item-to-item embeddings, etc., are stored on the cloud server. We name these features as \textbf{"centralized features"}. We find that the existing FL frameworks do not fully utilize these two sets of features. The main reason is that the federated and centralized features have different feature spaces, dimensions, and accessibility, and it is therefore challenging to utilize the federated features to improve the global model's performance. Also, it is impractical to move the centralized features to the edge since such features are large in size and the edge devices have limited storage and computation resources. 

The conventional centralized learning paradigm and the current FL paradigm only utilize one-sided cloud training or edge training. However, we reckon the edge-cloud collaboration is quite important in practice to have both better-generalized representations and more personalized user services \cite{yao2021device,gong2020edgerec,qian2022intelligent}. Thus, in this paper, we propose a \textbf{E}dge-\textbf{C}loud \textbf{C}ollaborative Knowledge \textbf{T}ransfer Framework (\textbf{ECCT}). In this framework, we train a big model on the server and a light weighted model on each edge device. The feature embeddings and prediction logits are shared between the edge and cloud to support knowledge transfer and collaboration. The edge and cloud models then adopt an alternating minimization (AM) approach \cite{he2020group,bolte2014proximal,attouch2010proximal} 
to utilize the transferred knowledge via embedding fusion and knowledge distillation. 

ECCT offers several advantages over conventional FL frameworks. Firstly, it allows for boosting local personalization by transferring knowledge from centralized features to on-device models. This improves personalization compared to FL methods. Secondly, ECCT enables model heterogeneity, allowing edge models to differ in architecture while still being compatible. Thirdly, ECCT naturally supports asynchronous training and is more robust when a partial selection of training devices exists. Finally, ECCT relieves communication burdens because transferred knowledge has a much smaller size than model parameters.

Our contributions are as follows.
\begin{itemize}[leftmargin=*]
    \item We develop the ECCT framework that jointly utilizes the cloud's centralized features and the edge's federated features. To the best of our knowledge, this is the first paper to use the two kinds of features to realize edge-cloud collaboration, and it has promising industrial applications.
    \item In ECCT, we implement bi-directional knowledge transfers between the edge and cloud by embedding fusion and knowledge distillation.
    \item Extensive experiments under both public and industrial datasets demonstrate the effectiveness of ECCT and show its broad and practical prospects in the industry.
\end{itemize}

\section{Related Works}
\textbf{Edge-cloud collaborative federated learning.} FedGKT \cite{he2020group} incorporates split learning in FL to realize edge-cloud collaboration. It trains a larger CNN model on the server based on the embeddings and logits from the devices. However, it does not utilize centralized data, and the knowledge from the cloud to the edge is weak by just transferring logits. There are works that take the cloud-side data as a public dataset to aid edge training via knowledge distillation \cite{zhang2022fedduap,cheng2021fedgems,nguyen2022cdkt}. We reckon it is not realistic to store such a public dataset at the edge devices, which hinders their applications in the industry.

\noindent\textbf{Edge-cloud collaborative recommender systems.} In \cite{yao2021device}, MoMoDistill is proposed to finetune the meta patches of the cloud RS model at the edge, and it can realize user personalization. In \cite{yao2022device}, a causal meta controller is trained to manage the tradeoff between the on-device and the cloud-based recommendations. We note that \cite{yao2021device,yao2022device} assume the centralized data is the whole sum of the federated data. But in realistic scenarios, the centralized and federated data have different features due to privacy and real-time issues. There are methods solving the inference problems in edge recommendation in terms of re-ranking \cite{gong2020edgerec} and item request \cite{qian2022intelligent}. 

\section{Methodology}
In Section \ref{subsect:formulation}, we give the main formulation of the proposed framework, while in Section \ref{subsect:modules}, we provide several practical add-ons that can be applied with the framework to further strengthen performance or privacy guarantees.
\subsection{Main Formulation} \label{subsect:formulation}
\noindent\textbf{Problem Setup. }
There are $K$ edge devices in the network. Specifically, the learning task is supervised learning for classification with $C$ categories in the entire dataset, which consists of two parts of features, geographically located at the server (centralized features, $\D$) and the devices (federated features, $\hat{\D}$). The $k$-th device has its own dataset of private and real-time features $\hat{\D}^k \coloneqq \left\{ \left( \hat{\X}_i^k, y^k_i \right)\right\}_{i=1}^{N^k}$, where $\hat{\X}_i$ is the $i$-th training sample, $y_i$ is the corresponding label of $\hat{\X}_i, y^k_i \in \{1,2,\ldots C\}$, and $N^k$ is the sample size of dataset $\hat{D}^k$ (also $D^k$). Thus, we have $\hat{\D}=\{\hat{\D}^1, \hat{\D}^2,\ldots,\hat{\D}^K\}$, $N=\sum_{k=1}^K N^k$. The server has all non-private and historical features of all devices as ${\D}^k \coloneqq \left\{ \left( {\X}_i^k, y^k_i \right)\right\}_{i=1}^{N^k}$, ${\D}=\{{\D}^1, {\D}^2,\ldots,{\D}^K\}$. 

\noindent\textbf{Learning Objectives. }
For edge device $k$, we deploy a small feature encoder (extractor) $\mathcal{E}_d^k$ and a small classifier $\mathcal{C}_d^k$, while for the cloud server, we deploy a large encoder $\mathcal{E}_s$ and a large-scale downstream model $\mathcal{C}_s$ (including the classifier). Let $\mathcal{W}_d^k = \{\mathcal{E}_d^k,\mathcal{C}_d^k\}$ and $\mathcal{W}_s = \{\mathcal{E}_s, \mathcal{C}_s\}$. Due to the separated features at the cloud and edge, we reformulate a single global model optimization into a non-convex optimization problem that requires us to train the server model $\mathcal{W}_s$ and the device model $\mathcal{W}_d^k$ simultaneously. The learning objectives of the server and device $k$ with respective loss functions of $\ell_s$ and $\ell_d^k$ are as follows 
\begin{align} \label{equ1}
\arg\min_{\mathcal{W}_s} F_s(\mathcal{W}_s) &= \arg\min_{\mathcal{W}_s} \sum_{k=1}^K \sum_{i=1}^{N^k}\ell_s\left(\mathcal{C}_s(h_{d,i}^k \oplus h_{s,i}^k), y_i^k\right),\\ \label{equ2}
\arg\min_{\mathcal{W}_d^k} F_d^k(\mathcal{W}_d^k) &= \arg\min_{\mathcal{W}_d^k}\sum_{i=1}^{N^k}\ell_d^k\left(\mathcal{C}_d^k(h_{d,i}^k \oplus h_{s,i}^k ), y_i^k\right),\\\label{equ3}
\text{where }h_{d,i}^k &= \mathcal{E}_d^k(\hat{\X}_i^k), ~h_{s,i}^k = \mathcal{E}_s(\X_i^k).
\end{align}
In Eqs. \ref{equ1} and \ref{equ2}, $h_{d,i}^k$ ($h_{s,i}^k$) refers to the extracted feature embedding of a sample $\hat{\X}_i^k$ ($\X_i^k$) using the device's (server's) encoder $\mathcal{E}_d^k$ ($\mathcal{E}_s$) and $\oplus$ denotes the concatenation operation that concatenates two embeddings into one. We optimize the above objectives via alternating minimization (AM) approach \cite{he2020group,bolte2014proximal,attouch2010proximal}. For the server's optimization, we fix the devices' embeddings $h_{d,i}^k$, and for device $k$'s optimization, we fix the server's embedding $h_{s,i}^k$. 

Apart from the embeddings, we also transfer the prediction logits and adopt knowledge distillation (KD) \cite{hinton2015distilling,lin2020ensemble} to strengthen knowledge learning. The shared embeddings and logits realize a bi-directional knowledge transfer in the edge-cloud collaborative training, and we adopt the loss functions for the server and edge device $k$ respectively as follows
\begin{align} \label{equ4}
    \ell_s = \ell_{CE} + \alpha_s\sum_{k=1}^K\ell_{KD}\left(z_d^k, z_s^k \right),\\
    \label{equ5}
    ~\ell_d^k = \ell_{CE}^k + \alpha_d\ell_{KD}^k\left(z_s^k, z_d^k \right),
\end{align}
where $\ell_{CE}$ is the cross-entropy loss between the predicted values and the ground truth labels, $\ell_{KD}$ is the Kullback Leibler Divergence function for knowledge distillation, and $z_s^k$ and $z_d^k$ are prediction logits. Besides, $\alpha_s$ and $\alpha_d$ are the hyper-parameters to control the strengths of knowledge distillation.

\begin{figure}[t]
\centering
\includegraphics[width=1.0\columnwidth]{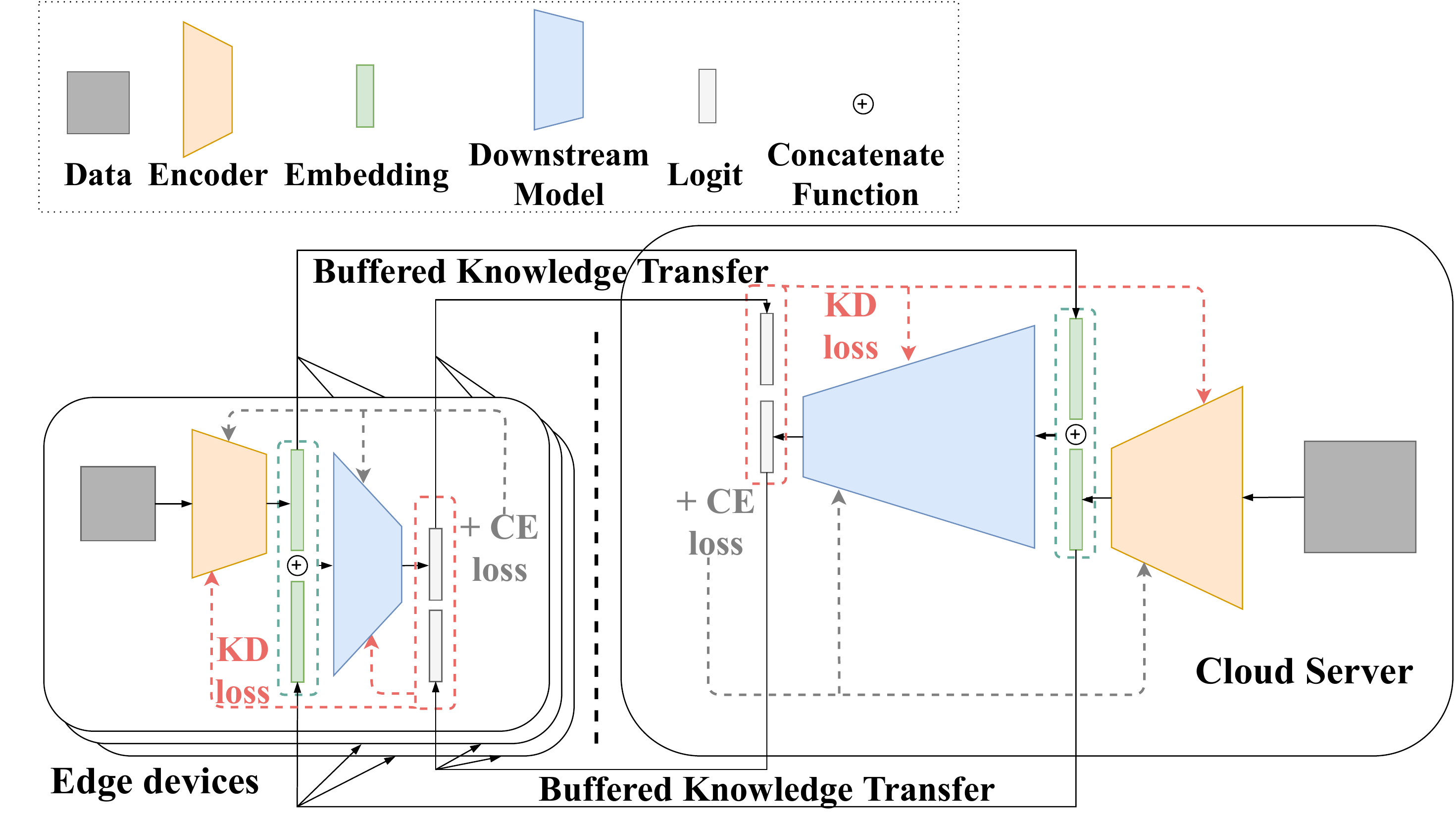}
\vspace{-0.6cm}
\caption{The proposed edge-cloud collaborative knowledge transfer framework.}
\label{fig:framework}
\vspace{-0.6cm}
\end{figure}

\noindent\textbf{Training Workflow. }
The demonstration of the proposed framework is shown in Figure \ref{fig:framework}. There are mainly three iterative steps in our edge-cloud collaborative learning framework: edge training, cloud training, and edge-cloud collaboration. 
(1)  \textit{Cloud training.} The cloud server conducts cloud training as that of Eq. \ref{equ4} for $E_s$ epochs. 
(2) \textit{Edge training.} The edge device $k$ conducts edge training as that of Eq. \ref{equ5} for $E_d^k$ epochs.
(3) \textit{Edge-cloud communication.} The edge device (cloud server) infers the embeddings via the encoder and the logits via the classifier on its local features and sends the embeddings and logits to the cloud server (edge devices).

Steps 1 and 2 can be conducted at the cloud and edge in parallel, and the AM-based method makes our framework more tolerant to asynchronous updates and communications. In practice, the device model accumulates its embeddings in a buffer, and when the buffer is full, it communicates with the server for transfer, and we name it buffered knowledge transfer. Each slice of buffered knowledge does not have to be generated from the same edge model. It relaxes the FL's synchronization requirements that the devices need to have the same model version (i.e. same edge-cloud communication frequency) and the same local epochs ($\forall i, j \in [K],~ E_d^i=E_d^j=E_d$). 
Moreover, due to the powerful computation resource in the cloud, we can have a larger number of cloud training epochs ($E_s \gg E_d$), and it can generate more knowledge-representative embeddings from enriched on-cloud features to aid the devices' models for fast training and better personalization. 

\noindent\textbf{Inference Strategy. }
During inference, there are two choices of strategies, i.e., the cloud-based and the edge-based ones. The cloud (edge) models use the transferred embeddings from the edge (cloud) and their own features to infer prediction for a given sample. Generally, the edge-based inference is more real-time and personalized while the cloud-based one is more generalized and robust. The inference strategies can be flexibly chosen according to the specific application scenarios.

\subsection{Practical Add-ons} \label{subsect:modules}
\noindent\textbf{Two-stage strategy.} We notice that in the early training phase when the models are less generalized, the knowledge of the logits is poor in representational power. As a result, the knowledge distillation may cause distortion in training. Thus, we only transfer the embeddings and use CE loss (i.e. $\alpha_s = \alpha_d$ = 0) in the first few training epochs, and additionally transfer the logits and use both KD and CE losses in later training phases.

\noindent\textbf{Filtered Knowledge Transfer.} We find that if a model predicts for some samples with wrong labels, the model is hardly generalized on these samples; therefore, taking the logits of these samples as knowledge will degrade performance in distillation \cite{cheng2021fedgems}. Thus, we adopt filtered knowledge transfer that only uses the right logits for knowledge distillation.

\noindent\textbf{Privacy Guarantee.} The transfer of embeddings bears less privacy risk than directly transferring the device's raw private features \cite{chen2021fedgraph,fu2022federated}. In our framework, the privacy guarantee can be further strengthened by applying differential privacy methods on the embeddings from the devices \cite{abadi2016deep,wei2020federated}.

\vspace{-0.12cm}
\section{Experiments}
 In the experiments, we evaluate our framework on three datasets: CIFAR10, Avazu, and IndustryData. We introduce experimental settings in Section \ref{subsect:exp_setting}, present the results on CIFAR10 with synthetic feature split in Section \ref{subsect:exp_cifar10}, and illustrate the results on RS datasets (Avazu and IndustryData) in Section \ref{subsect:rs}.

\vspace{-0.1cm}
\subsection{Experimental Settings} \label{subsect:exp_setting}
For fair comparisons, we consider three feature settings: "F" uses only federated features, "C2F" moves \textbf{c}entralized features \textbf{to} the edge as \textbf{f}ederated features, and "C\&F" combines centralized features at the cloud and federated features at the edge. We note that "C2F" is not practical in real-world scenarios due to storage limitations and communication overhead. However, we include it for comprehensive evaluation. Our experiments evaluate personalization results on local test sets of edge devices, using three datasets. For the first 50 rounds of training, $\alpha_s$ and $\alpha_d$ are both set to 0, and then set to 1 for the latter 50 rounds. The transferred embeddings have a size of 128, and we employ an MLP with two layers as the downstream classifier.

\noindent\textbf{Datasets.} \textbf{CIFAR10} \cite{krizhevsky2009learning} is an image dataset for classification, and we conduct synthetic feature split to generate federated and centralized features. Specifically, given a 32×32 image, we split it into a 10×32 image at the edge and a 22×32 image at the cloud. We deploy ResNet-8 at the edge and ResNet-56 at the cloud as the encoders. The number of devices is 20. \textbf{Avazu} \cite{avazu-ctr-prediction} is a public dataset for click-through rate (CTR) prediction in real-world RS. We randomly sample 200,000 data samples and assign them to each device by device\_ip, and there are 1769 devices. We set the sparse features, which are mostly about the item details (e.g. app\_category), as the cloud-side centralized features and assume the dense features as the federated features. We use AutoInt \cite{song2019autoint} as the backbone encoders. \textbf{IndustryData} is a real-world industrial dataset for RS, extracted from system logs from Alipay APP. We randomly sample 500,000 data samples and assign them to 2000 devices. We allocate the user's sensitive features at the edge, like u\_gender and u\_occupation, and the historical interactive features as the centralized features at the cloud. AutoInt is used as the encoder, and we conduct experiments on both CTR and conversion rate (CVR) prediction tasks.

\noindent\textbf{Baselines.} We compare our framework with FL baselines, mainly FedAvg \cite{mcmahan2017communication} (the general FL framework) and FedGKT \cite{he2020group} (the SOTA edge-cloud collaborative FL framework). For the experiments with CIFAR10, since the split features have the same feature spaces, we also test distillation-based FL methods (FedDF \cite{lin2020ensemble} and FedBE \cite{chen2020fedbe}) that can utilize the centralized features via distillation. 

\subsection{Synthetic Feature Split} \label{subsect:exp_cifar10}

In this section, experiments are conducted on CIFAR10 with synthetic feature split. 
Table \ref{tab:cifar10_noniid} shows the results under different device data heterogeneity and numbers of edge training epochs ($E_d$) using Dirichlet sampling \cite{lin2020ensemble} to assign devices with NonIID samples. Our method outperforms the baselines in both IID and NonIID settings. Moreover, our method shows a smaller decrease in performance compared with FedGKT when device model heterogeneity exists. Distillation (FedDF and FedBE) on the centralized features results in worse model performance, indicating that conventional FL methods cannot effectively utilize centralized features. 

Table \ref{tab:cifar10_asyn} presents the results under asynchronous settings. The degradation in performance is stronger in FedAvg than in ECCT, especially when combined with uniformly random partial device selection in each epoch of training. 

Table \ref{tab:cifar10_ratio} showcases the performance of different methods with various device numbers and selection ratios. Our ECCT outperforms FedAvg and FedGKT under all settings and is more tolerant with different participation of edge devices.




\begin{table}[t] \small
\caption{Top-1 test accuracy (\%) results on CIFAR10 under different device data heterogeneity and $E_d$. "homo"/"hetero": device model homogeneity/heterogeneity.}
\vspace{-0.4cm}
\centering
\begin{tabular}{c|c|c|c|c|c}
\toprule 
&Data heterogeneity&\multicolumn{2}{c}{IID}&\multicolumn{2}{c}{NonIID}\\ 
\midrule 
Feature&Method/$E_d$ &1&3&1&3\\
\midrule
\multirow{3}{*}{F}&FedAvg &57.36&59.75&{52.36}&{56.96}\\
&FedGKT (homo)&{66.78}&{67.90}&38.88&44.59\\
&FedGKT (hetero)&51.57&54.34&23.36&26.20\\
\midrule
\multirow{3}{*}{C2F}&FedAvg&61.81&66.32&{57.09}&{62.69}\\
&FedGKT (homo)&{78.44}&{79.05}&30.91&57.70\\
&FedGKT (hetero)&66.09&66.04&30.38&49.30\\
\midrule
\multirow{4}{*}{C\&F}&FedDF&31.9&49.14&28.86&41.36\\
&FedBE&38.28&51.55&31.73&43.09\\
&\textbf{ECCT (homo)}&\textbf{81.89}&\textbf{82.46}&\textbf{63.06}&\textbf{62.61}\\
&\textbf{ECCT (hetero)}&\textbf{77.31}&\textbf{77.18}&\textbf{60.86}&\textbf{64.06}\\
\bottomrule
\end{tabular}
\label{tab:cifar10_noniid}
\vspace{-0.2cm}
\end{table}

\begin{table}[t]  \small
\caption{Top-1 test accuracy (\%) results on CIFAR10 under asynchronization.}
\centering
\vspace{-0.4cm}
\begin{tabular}{c|cc|cc}
\toprule
Method& \multicolumn{2}{c}{FedAvg} &\multicolumn{2}{c}{\textbf{ECCT}}\\
\midrule 
Select ratio &1.0 &0.5 &1.0 &0.5\\
\midrule
\multirow{2}{*}{Syn.}&62.69&55.78 &67.81	&69.28\\
&- &{$11.0\%\downarrow$}	&- &{$2.1\%\uparrow$}	\\
\midrule
\multirow{2}{*}{Asyn. version}& 60.51&53.29 &67.56	&64.42\\
&{$3.5\%\downarrow$}	&{$15.0\%\downarrow$} &{$0.4\%\downarrow$}	&{$5.3\%\downarrow$}	\\
\midrule
\multirow{2}{*}{Asyn. epoch}& 59.55&50.93 &66.92	&64.65\\
&{$5.0\%\downarrow$}	&{$18.8\%\downarrow$} &{$1.3\%\downarrow$}	&{$4.9\%\downarrow$}\\
\midrule
\multirow{2}{*}{Asyn. both}& 58.47&51.49 &67.31	&64.81\\
&{$6.7\%\downarrow$}	&{$17.9\%\downarrow$}&{$0.7\%\downarrow$}	&{$4.6\%\downarrow$}\\
\bottomrule
\end{tabular}
\label{tab:cifar10_asyn}
\vspace{-0.45cm}
\end{table}

\begin{table}[t]  \small
\caption{Top-1 test accuracy (\%) results on CIFAR10 under different device numbers and device selection ratios.}
\vspace{-0.4cm}
\centering
\begin{tabular}{c|ccc|ccc}
\toprule
Device Num.&\multicolumn{3}{c}{50}&\multicolumn{3}{c}{100}\\
\midrule 
Select ratio &0.6 &0.3 &0.1 &0.6 &0.3 &0.1 \\
\midrule
FedAvg  &47.64	&33.99	&14.12	&44.04	&31.55	&16.19\\
FedGKT	&31.06	&44.84	&45.16	&59.68	&57.82	&52.53\\
\midrule
\textbf{ECCT}	&\textbf{68.43}	&\textbf{68.68}	&\textbf{45.33}	&\textbf{78.36}	&\textbf{75.91}	&\textbf{67.01}\\
\bottomrule
\end{tabular}
\label{tab:cifar10_ratio}
\vspace{-0.2cm}
\end{table}

\subsection{Recommender Systems} \label{subsect:rs}
In this section, we implement the methods in RS, and both public (Avazu) and industrial datasets (IndustryData) are used. 

Firstly, we conduct experiments on Avazu for CTR prediction tasks and demonstrate the results in Table \ref{tab:avazu}. Our method is superior to FedAvg and FedGKT in terms of AUC, even moving all the centralized features to the edge ("C2F"). Secondly, we test the methods on IndustryData for CTR and CVR prediction tasks. The results in Table \ref{tab:alipay} illustrate that our edge-cloud collaborative framework performs better than the FL methods, especially in the CVR task. It shows that the framework has large potential in the industry, especially for online recommendations.

We notice that FedGKT has poor results in RS, and we find it reasonable. In FedGKT, the knowledge from the cloud to the edge is solely based on the logits. For binary classification tasks like predicting CTR and CVR, the logits contain no more information than the labels, so FedGKT fails to make an effective transfer from the cloud to the edge. However, our framework incorporates more knowledge in the transferred embeddings.

\begin{table}[t] \small
\caption{CTR prediction for Avazu.}
\vspace{-0.4cm}
\centering
\begin{tabular}{c|cc|cc|c}
\toprule
Feature& \multicolumn{2}{c}{F}& \multicolumn{2}{c}{C2F}&C\&F\\
\midrule
Method&FedAvg&FedGKT&FedAvg&FedGKT&\textbf{ECCT}\\
\midrule 
AUC & 0.5783& 	0.5293& 	0.6595& 	0.5246& 	\textbf{0.6694} \\
MSE&	0.1370& 	0.1488& 	0.1320& 	0.1559& 	\textbf{0.1373} \\
\bottomrule
\end{tabular}
\label{tab:avazu}
\vspace{-0.2cm}
\end{table}

\begin{table}[t] \small
\caption{CTR and CVR predictions for IndustryData.}
\vspace{-0.4cm}
\centering
\begin{tabular}{c|cc|cc}
\toprule
Task&\multicolumn{2}{c}{CTR}&\multicolumn{2}{c}{CVR}\\
\midrule 
Method/Metric&AUC &	MSE&	AUC &	MSE\\
\midrule 
FedAvg&	0.6829& 	0.0879& 	0.8145& 	0.0199 \\
FedGKT&	0.5409& 	0.1023& 	0.6068& 	0.0262 \\
\midrule
\textbf{ECCT}&	\textbf{0.6885}&	\textbf{0.0875}& 	\textbf{0.8754}& 	\textbf{0.0184} \\
\bottomrule
\end{tabular}
\label{tab:alipay}
\vspace{-0.22cm}
\end{table}

\section{Conclusion}
In this paper, we proposed an edge-cloud collaborative knowledge transfer framework to jointly utilize the centralized and federated features. The proposed method has several advantages over the previous federated learning methods: boosting edge personalization, enabling model heterogeneity, tolerating asynchronization, and relieving communication burdens between edge and cloud.

\begin{acks}
This work was supported by the National Key Research and Development Project of China (2021ZD0110400), the National Natural Science Foundation of China (U19B2042), the Program of Zhejiang Province Science and Technology (2022C01044), The University Synergy Innovation Program of Anhui Province (GXXT-2021-004), Academy Of Social Governance Zhejiang University, Fundamental Research Funds for the Central Universities (226-2022-00064).
\end{acks}

\newpage
\bibliographystyle{ACM-Reference-Format}
\balance
\bibliography{main-sigir2023}

\end{document}